\documentclass[preprint, review, 12pt]{elsarticle}

\usepackage{graphicx}
\usepackage{amssymb}

\usepackage{amsmath}

\newcommand{\cP}{\mathcal{P}}

\begin{document}

\begin{frontmatter}


\title{Should You Derive, Or Let the Data Drive?\\ An Optimization Framework for \\ Hybrid First-Principles Data-Driven Modeling}



\author[labelMIT]{R. R. Lam\corref{cor}}
\ead{rlam@mit.edu}
\author[labelIBM]{L. Horesh}
\ead{lhoresh@us.ibm.com}
\author[labelTAU]{H. Avron}
\ead{haimav@post.tau.ac.il}
\author[labelMIT]{K. E. Willcox}
\ead{kwillcox@mit.edu}
\address[labelMIT]{Massachusetts Institute of Technology, Cambridge, Massachusetts}
\address[labelIBM]{IBM Research, Yorktown Heights, New-York}
\address[labelTAU]{Tel Aviv University, Tel Aviv, Israel}
\cortext[cor]{Corresponding author.}

\begin{abstract}
Mathematical models are used extensively for diverse tasks including analysis, optimization, and decision making.
Frequently, those models are principled but imperfect representations of reality.
This is either due to incomplete physical description of the underlying phenomenon (simplified governing equations, defective boundary conditions, etc.), or due to numerical approximations (discretization, linearization, round-off error, etc.).
Model misspecification can lead to erroneous model predictions, and respectively suboptimal decisions associated with the intended end-goal task.
To mitigate this effect, one can amend the available model using limited data produced by experiments or higher fidelity models.
A large body of research has focused on estimating explicit model parameters.
This work takes a different perspective and targets the construction of a correction model operator with implicit attributes.
We investigate the case where the end-goal is inversion and illustrate how appropriate choices of properties imposed upon the correction and corrected operator lead to improved end-goal insights.
\end{abstract}




\end{frontmatter}



\section{Introduction} 
\label{sub:introduction}

	Mathematical models play a central role in numerous fields: description, analysis, optimization, inversion, decision making, etc.
	Development of a model requires proper balance between multiple objectives that influence the model realism and its complexity.
	One desirable property is to achieve high fidelity: an accurate representation of the phenomenon of interest.
	Another objective is to develop a model that is tractable to solve either analytically or numerically.
	These competing requirements typically lead to a trade-off between fidelity and complexity of the model.
    For example, one might choose to use a simplified set of governing equations, sacrificing fidelity in favor of reduced computational complexity.
    As a consequence, models often form imperfect representations of the true physical process,
	and thus, also impact and influence the end-goal task.
	In this paper, we refer to such imperfect models as \textit{misspecified models}.
	The inadequacies associated with such models are often disregarded, or over-simplistically treated as noise.

	For the sake of clarity, we  define the operators and their respective mappings.
	We describe the process or phenomenon under investigation by $T$, which is characterized by a set of model parameters $m$, control terms $q$, and experimental design $\theta$.
	The potential set of state parameters $u_T$ is given by the following equation:
	\begin{eqnarray}
		T(u_T,q,m;\theta) = 0. \label{eq:T_definition}
	\end{eqnarray}
	For simplicity, we assume that such $u_T$ exists everywhere and is unique,
	and denote with $P_\theta$ the observation operator, which links a state $u_T$ to the observations $d$, and is defined by:
	\begin{equation}
		d = P_\theta (u_T) + \epsilon(\theta).\label{eq:d_definition}
	\end{equation}
	In the above, we also incorporate a noise term, $\epsilon$, to account for intrinsically stochastic error, which is independent of the process.

	Rarely, a mathematical model describes conclusively
	the underlying phenomenon or process, $T$; thus, in practice, often one resorts to use of an imperfect model $M$
 	to describe the behavior of the system under investigation.
	Since the model $M$ is misspecified, we shall denote by $R$ the residual due to model discrepancy:
	\begin{eqnarray}
		\underbrace{R(u,q,m;\theta)}_{{\textrm {model discrepancy}}} = \underbrace{T(u,q,m;\theta)}_{\textrm{fully specified model}} - \underbrace{M(u,q,m;\theta)}_{\textrm{misspecified model}}.
	\end{eqnarray}
	The use of a misspecified model $M$, instead of of the fully specified one, $T$, leads to an inaccurate state parameter $u_M$ (instead of $u_T$) satisfying:
	\begin{align}
		M(u_M,q,m;\theta) = 0. \label{eq:M_definition}
	\end{align}
	Revisiting the observation $d$, this impairs the predicted data integrity through introduction of the observation error term $\xi$:
	\begin{eqnarray}
		&d = \underbrace{P_\theta(u_M)}_{\text{misspecified prediction}} + \ \underbrace{\xi(u_{M},u_{T},q,m;\theta) + \epsilon(\theta)}_{\zeta(u_{M},u_{T},q,m;\theta)}.
		\label{eq:mis_eq}
	\end{eqnarray}
	In conventional settings, both $\xi$ and $\epsilon$, aggregated above as $\zeta$, are commonly regarded
	as an apparent ``noise''.
	Yet, as the term noise in fact hosts various factors of different characteristics, we ought to be more careful with the terminology at this point.
	We therefore decouple the ``noise'' term, $\zeta$, into two components: the first, $\xi(u_{M},u_{T},q,m;\theta) $ depends upon  the parameters $m$, controls $q$ and experimental design $\theta$, whereas the latter, $\epsilon(\theta)$ depends solely upon the experimental design.

	Given this setup, it is
 	possible to decouple the effect of instrumentation noise and intrinsic stochastic effects, from misspecification of the
	model.
	This separation may seem artificial, as in principle,  instrumentation error can be also included in a more comprehensive model that includes the measurement apparatus in addition to the system under investigation.
	Yet, in the context of this study, such separation already provides another level of granularity overlooked
	in the traditional setup.
		
	Following the above notational exposition, it is now possible to distinguish between two complementary challenges:
	\begin{itemize}
		\item  Model {\it reduction} - aims at reducing a model complexity, potentially at the expense of a decreased fidelity: \sloppy$|\zeta_{r}(u_{M},u_{T},q, m; \theta)| \ge |{\zeta}(u_{M},u_{T},q, m; \theta)|$.
		\item  Model {\it improvement} - aims at increasing a model fidelity $|\zeta_{i}(u_{M},u_{T},q, m; \theta)| \le |{\zeta}(u_{M},u_{T},q, m; \theta)|$, potentially at the expense of increased complexity.
	\end{itemize}
	While the former has been investigated extensively
	(see e.g., \cite{antoulas2005approximation, benner2015survey, Patera_book}),
	little attention has been given to the development of consistent, derivable frameworks for the latter challenge.

	The goal of this study is to propose a framework for inference of a {\it correction} model, $C$, for the misspecified term, $M$, such that the {\it corrected}\footnote{Note that we distinguish between the correction model and the corrected model, as different requirements may be imposed upon each of them.} model
	$M(u,q,m;\theta)+C(u,q,m;\theta)$
	performs better for a desired end-goal task, in a sense yet to be defined. Thus, paraphrasing George Box's statement ``All models are wrong, but some are useful'', our goal is to proactively make models more useful, through learning an approximated correction model for the misspecified elements.
	The framework is particularly suited for situations where a first-principles model provides an inadequate approximation of the underlying phenomenon or process.
	Yet, as not all governing aspects of the phenomenon are fully realized in the model (either due to lack of knowledge, understanding or computational limitations), a hybrid first-principles data-driven approach, in which the cardinal missing elements of the model are learned from data, can enhance the fidelity and utility of the model.

	In Table~\ref{table:first-vs-driven}, some of the common characteristics of first-principles modeling are compared against those of data-driven approaches.
	Rather than choosing between the two, the approach taken in this study is to combine the strengths of the two approaches together and thereby attain models of superior attributes for a given task at hand.
	
	\begin{table}
		\centering
		\footnotesize
		\resizebox{\textwidth}{!}{%
		\begin{tabular}{|| p{2.5cm} | | p{4.5cm}  || p{4.5cm}| |}
			\hline
			Attribute & Data-driven & First-Principles  \\ \hline
			Adaptability \&  deployability  & + Generic frameworks, easily adapted to new problems  & - Requires complex derivation, specific to the application  \\ \hline
			Domain expertise reliance  & + Provide useful results with little domain knowledge & - Depends heavily upon domain expertise  \\ \hline
			Data availability reliance  & - Usually requires Big Data to train upon & + Usually can be derived from Small Data \\ \hline
			Fidelity \& robustness  &  - Limited generality associated with the training set span and complexity & + Universal links of complex relations \\ \hline
			Interpretability  &  - Limited due to functional form rigidity & + Physically meaningful and interpretable link between parameters \\  \hline
		\end{tabular}
	}
	\caption{Common characteristics of first-principles modeling and data-driven approaches.}
	\label{table:first-vs-driven}
	\end{table}

	\vspace{5 mm}

	This paper is organized as follows.
	Section \ref{sec:priorart} reviews some of the prior art in the area. 	
	In Section \ref{sec:problem_definition}, we formally define the problem, the involved spaces, and properties of the correction and corrected models.
	In Section~\ref{sub:model_choices_and_simplifications}, for the sake of illustration, we describe a particular instance of the general framework.
	In Section \ref{sub:minimization_problem}, we present several simplifications of the optimization problem to keep it tractable.
	A numerical study is presented in Section \ref{sec:numerical_study_airfoil_interactions}, to demonstrate the utility of the framework.
	Finally, we close with conclusions and perspectives in Section \ref{sub:conclusion}.

\section{Prior Art} 
\label{sec:priorart}

	A great body of research has been dedicated, somewhat disjointly, to either first-principles modeling \cite{newton1833philosophiae,maxwell1881treatise}, or statistical modeling \cite{coles2001introduction,breiman2001statistical,freedman2009statistical}.
	A few interesting instances in which model fidelity is enhanced through a combination of the two can be found in the literature.
	There are several types of discrepancy that reduce model fidelity.
	Kennedy and O'Hagan \cite{kennedy2001bayesian} define them, including model inadequacy: the ``difference between the true mean value of a real world process and the code value at the true value of the inputs''.
	They mitigate this discrepancy using Gaussian Process models to approximate the difference between a low-fidelity model output and high-fidelity data \cite{kennedy2000predicting}.
	Recent work \cite{oliver2011bayesian,oliver2014formulation} has tackled the inadequacy issue in fully-developed channel flows and wall-bounded flows modeling by using a stochastic inadequacy model and calibrating the uncertain parameters with available high-fidelity data.

	Other studies that leverage data to improve a model include \cite{hamann2015multi},
	which considers input-output pairs of multiple physical models as a training set for a machine learning framework. This is used to create a blended model of superior fidelity for each original model.
	Zhang's thesis~\cite{zhang2008improved} focuses on how the collection of data influence modeling, and develops a dynamic Design of Experiments framework.
	There is an interesting interplay between experimental design \cite{haber2008numerical,haber2010numerical} and model correction.
	Both may modify the observation operator, in order to better reflect how observables and model parameters are related.
	However, in experimental design, a common assumption is that the underlying phenomenon is understood,
	and the challenge is to gain more understanding of the model parameters.
	In contrast, model correction challenges the fundamental assumption that the governing behavior of the system is known.

	Correcting a model has also been studied in the context of convolution operators,
	where a notable body of research is devoted to estimation of impulse response functions based on observational data in the form of blind-deconvolution \cite{bell1995information,ayers1988iterative,levin2009understanding}.

	A stochastic optimization framework for linear model correction has been proposed in \cite{hao2014nuclear,kilmer2014householder}.
	The approach provides means to balance between fidelity of the corrected model and complexity of a non-parameterized correction.
	Since often there are also properties that one may wish to attribute to the corrected model, rather than to the correction itself, in this study, we generalize the framework to allow greater control upon the designed corrected model.


\section{Mathematical Framework} 
\label{sec:problem_definition}

\subsection{Variables and Spaces Definition} 
\label{sub:variables_and_spaces_definition}

	In the following, we consider the simplified situation where the model parameter $m$ and the experimental design setup $\theta$ are fixed. In this case, the fully specified model $T$ of Eq.~\ref{eq:T_definition} becomes:
	\begin{align}
		T(u_{T},q)=0,
	\end{align}
	and the misspecified model $M$ of Eq.~\ref{eq:M_definition} is:
	\begin{align}
		M(u_{M},q)=0.\label{eq:M_implicit}
	\end{align}

	The observation model  Eq.~\ref{eq:d_definition} becomes:
	\begin{align}
		d = P(u_{T})+\epsilon,
	\end{align}
	where $\epsilon$ represents the noise. Note that $d$ (which could be multidimensional) depends on the control $q$ through the state\footnote{Note that the state $u_T$ is in general not directly observable.} $u_T$, defined by $T(u_{T},q)=0$.
	We consider the case where we have $N_T$ observations. Each observation $d^{(i)}$, $\ i=1,\ldots,N_T$ corresponds to a particular realization of the control $q^{(i)}$, $i=1,\ldots,N_T$. 
	The collection $Q$ of such controls and the collection $D$ of associated observations define a training set.

	We now consider the problem of estimating a correction model $C$, such that the corrected model
	behaves like the fully specified one $T$ for a specific end-goal task.

\subsection{Properties of the Correction and Corrected Operator} 
\label{sub:properties_of_the_correction_corrected_operator}

	Given a misspecified model and a training set from the fully specified model, our goal is to form a correction model $C$, such that the corrected model achieves a higher \textit{fidelity} than the misspecified one for an end-goal task.
	One way to achieve this goal is to design $C$ such that the discrepancy on the training set is minimized. When little data regarding the process of interest is available,
	such a strategy alone is problematic because the problem may be under-determined.
	
	Depending on the end-goal objective, some properties, or \textit{virtues}, of the corrected model may be desirable.
	Here, we define a virtue to be a mapping, $\mathcal{V}$, from the misspecified and correction model to a real valued scalar.
	Possible virtues could be computational speed (for a real-time application or scalability considerations), conditioning (for matrix inversion), stability (for control), etc.
	Enforcing such virtues restricts the space of admissible correction models $C$.
	
	Finally, we may want to impose some \textit{structure} upon the correction.
	Here, we define a structure to be a mapping, $\mathcal{S}$, from the correction model to a real valued scalar.
	This structure could be explicit: e.g., a parametric form of the correction, symmetry, tri-banded structure, invariance under some operations, etc.
	This structure could also be implicit: sparsity of the correction, low-rankness, existence in a low dimensional manifold, etc.
	Low dimensional structural preferences
	can be instrumental in ensuring that $C$ is `small' compared to $M$, as it is indeed a correction.
	
	We formulate the problem of estimating  the correction model as an optimization problem involving the three aforementioned quantities: fidelity, virtue, and structure:
	\begin{eqnarray}
		C=&\operatornamewithlimits{argmin}\limits_{C}~\mathcal{G}[\mathcal{F}(M,C,D,Q)]\label{eq:G}\\
		\text{s.t. }&\mathcal{V}(M,C)\leq\nu\\
		&\mathcal{S}(C)\leq\tau, \label{eq:S}
	\end{eqnarray}
	where $\mathcal{G}$ represents a noise model, $\mathcal{F}$ is the functional form that quantifies the fidelity of the corrected model with respect to the training set comprising a set of controls $Q$ and a set of corresponding observations $D$, $\mathcal{V}$ quantifies the virtue of the corrected operator, and $\mathcal{S}$ qualifies the structure of the correction $C$. 
	The  values
	$\nu \in \mathbb{R}$ and $\tau \in \mathbb{R}$ are user-specified constants that specify the desired restrictions on the virtue metric and the structure metric, respectively.
	The next section gives an illustrative example for choices of  these quantities. An important aspect of ongoing work is to identify particular choices for quantification of fidelity, virtue and structure that lead to a well-posed optimization problem Eq.~\ref{eq:G}--\ref{eq:S}.

\section{An Example of Model Choices} 
\label{sub:model_choices_and_simplifications}

	While the framework described in Section~\ref{sub:properties_of_the_correction_corrected_operator} is  general, we now make some model choices for the purpose of illustration.

	We consider the case where the misspecified model $M$ is characterized by an invertible linear operator.
	In this case, the general misspecified model Eq.~\ref{eq:M_implicit} is given
	by:
	\begin{align}
		M(\mathbf{u}_M,\mathbf{q})=\mathcal{M}\mathbf{u}_M-\mathbf{q}=0,
	\end{align}
	where $\mathcal{M}$ is 
	a square invertible matrix mapping 
	a state vector $\mathbf{u}_{M}$ to a control vector $\mathbf{q}$.
	We restrict our attention to the class of additive correction model $C$, characterized by linear 
	operators $\mathcal{C}$; thus, a corrected operator acting upon a state vector $\mathbf{u}$ can be written $(\mathcal{M}+\mathcal{C})\mathbf{u}$.

	The correction of a misspecified model is tailored to a given end-goal task.
	In this example, we choose this task to be the resolution of the inverse problem of estimating the state given the control $\mathbf{q}$.	
	We are given a training set of observed data and controls $\{D,Q\}$ from the fully specified model, where
	$D$ (respectively $Q$) is a matrix whose $i^{th}$ column is the $i^{th}$ observation vector $\mathbf{d}^{(i)}$ (respectively control vector $\mathbf{q}^{(i)}$) of the training set.
	We consider the case where the data observed is a subset of the state field, so the observation operator is represented
	by a linear sampling operator $\mathcal{P}$ (for instance a matrix with entries 0 and 1) that selects the corresponding observed part of the state.
	
	We now specify the three components of the optimization framework: fidelity, virtue, and structure. 
	Given a misspecified model $M$, a correction $C$, 
	and a training set $D,Q$, we define $\mathcal{F}$ to be the error in the observable data: 
	$\mathcal{F}(M,C,D,Q) = \mathcal{P}^{\top}[(\mathcal{M}+\mathcal{C})^{-1}Q]-D$.
	We also choose the noise model $\mathcal{G}$ to be the square of the Frobenius norm: $\mathcal{G}[\cdot]=\left\lVert \cdot\right\lVert_{F}^{2}$.

	We can now define the fidelity term, which we also refer to as the \textit{inverse error}, to be:
	\begin{eqnarray}
		\mathcal{G}[\mathcal{F}(M,C,D,Q)]=\left\lVert \mathcal{P}^{\top}[(\mathcal{M}+\mathcal{C})^{-1}Q]-D\right\lVert_{F}^{2}.\label{eq:def_fidelity}
	\end{eqnarray}
	Note that computing the inverse error requires the corrected model to be invertible.
	
	A natural virtue to impose upon the corrected model for the task of inversion is a low condition number for the corrected operator:
	\begin{align}
		\mathcal{V}(M,C)=\kappa(\mathcal{M}+\mathcal{C}).\label{eq:def_virtue}
	\end{align}
	A low condition number, $\kappa$,  avoids ill-conditioned inversion and numerical issues.
	In addition, the condition number quantifies the spread of the eigenvalues, which affects the convergence rate of iterative methods (e.g., GMRES).
	Thus, when dealing with large systems, a low condition number
	is likely to reduce the number of solver iterations or equally, to offer greater precision for a similar computational budget.
	
	For structure, we choose to impose a low-rank structure on the correction operator $\mathcal{C}$.
	This corresponds to ensuring that the correction is small.
	Thus we have,
	\begin{align}
		\mathcal{S}(C)=\text{rank}(\mathcal{C}).\label{eq:def_structure}
	\end{align}

	These are specific choices to demonstrate the proposed concepts of fidelity, virtue, and structure. Other choices are possible, as mentioned in Section~\ref{sub:properties_of_the_correction_corrected_operator}, depending on the context of the problem.



\section{Optimization Problem} 
\label{sub:minimization_problem}

	For the sake of tractability, we
 	now make additional simplifications
	to 
  	the optimization problem defined both by Eq.~\ref{eq:G} and the choices and assumptions of Section~\ref{sub:model_choices_and_simplifications} (Eq.~\ref{eq:def_fidelity}--\ref{eq:def_structure}).

	First, we convert the virtue constraint to a regularization term and write the optimization as:
	\begin{eqnarray}
		\mathcal{C} = &\operatornamewithlimits{argmin}\limits_{\mathcal{C}}~ \left\lVert \cP^{\top}[(\mathcal{M}+\mathcal{C})^{-1}Q]-D\right\lVert_{F}^{2} + \lambda \kappa(\mathcal{M}+\mathcal{C})\\
		\text{s.t. }&\text{rank}(\mathcal{C})\leq \tau,
	\end{eqnarray}		
	where $\lambda$ is a regularization parameter.

	We now simplify the formulation in two ways. First, we use the square of the Frobenius norm of the inverse of the corrected operator $\left\lVert(\mathcal{M}+\mathcal{C})^{-1}\right\lVert^{2}_{F}$  as a proxy for the condition number $\kappa(\mathcal{M}+\mathcal{C})$.
	This is motivated by the idea that the Frobenius norm of an inverse matrix $K^{-1}$ provides a lower bound on the smallest singular value $\sigma_{min}$ of $K$ as $\left\lVert K^{-1}\right\lVert_{F}^{2} \geq \sigma^{-2}_{min}$,
	and thus controls indirectly the condition number $\kappa(K)$.

	Second, the constraint on the rank is approximated by a constraint on the nuclear norm, which is the tightest convex relaxation of the rank operator.
	This leads to the following optimization problem:
	\begin{eqnarray}
		\mathcal{C} = &\operatornamewithlimits{argmin}\limits_{\mathcal{C}}~ \left\lVert \cP^{\top}[(\mathcal{M}+\mathcal{C})^{-1}Q]-D\right\lVert_{F}^{2} + \lambda \left\lVert(\mathcal{M}+\mathcal{C})^{-1}\right\lVert^{2}_{F}\label{eq:reformulated_optim}\\
		\text{s.t. }&\left\lVert \mathcal{C}\right\lVert_{*}\leq \delta,\label{eq:nuclear_constraint}
	\end{eqnarray}		
    where $\left\lVert \cdot\right\lVert_{*}$ denotes the nuclear norm and $\delta \in \mathbb{R}$ is the specified limit on the nuclear norm of $\mathcal{C}$. %

	The gradient of the objective function can be computed in closed form using the following equations, where we define $A=\mathcal{M}+\mathcal{C}$ for compactness:
	\begin{eqnarray}
		&\frac{\partial}{\partial \mathcal{C}}\left\lVert \cP^{\top}[A^{-1}Q]-D\right\lVert_{F}^{2} = -2A^{-\top}\cP[\cP^{\top}[A^{-1}Q]-D]Q^{\top}A^{-\top}\\
		&\frac{\partial}{\partial \mathcal{C}}\left\lVert A^{-1}\right\lVert_F^2 = -2A^{-\top}A^{-1}A^{-\top}.
	\end{eqnarray}
	The constraint Eq.~\ref{eq:nuclear_constraint} can be handled in various ways, such as semidefinite programming, projection and proximity, or Frank-Wolfe convex reformulation \cite{liu2009interior,toh2010accelerated,avron2012efficient,hazan2008sparse}.

\section{Numerical Study: Wings Interaction} 
\label{sec:numerical_study_airfoil_interactions}

	We proceed with a validation study with a simple illustrative problem in which the fully specified model $\mathcal{T}$ is known.
	
	\subsection{Problem setup}
\begin{figure}[!htbp]
	\centering
		\begin{minipage}[t]{0.47\textwidth}
			\includegraphics[width=\textwidth]{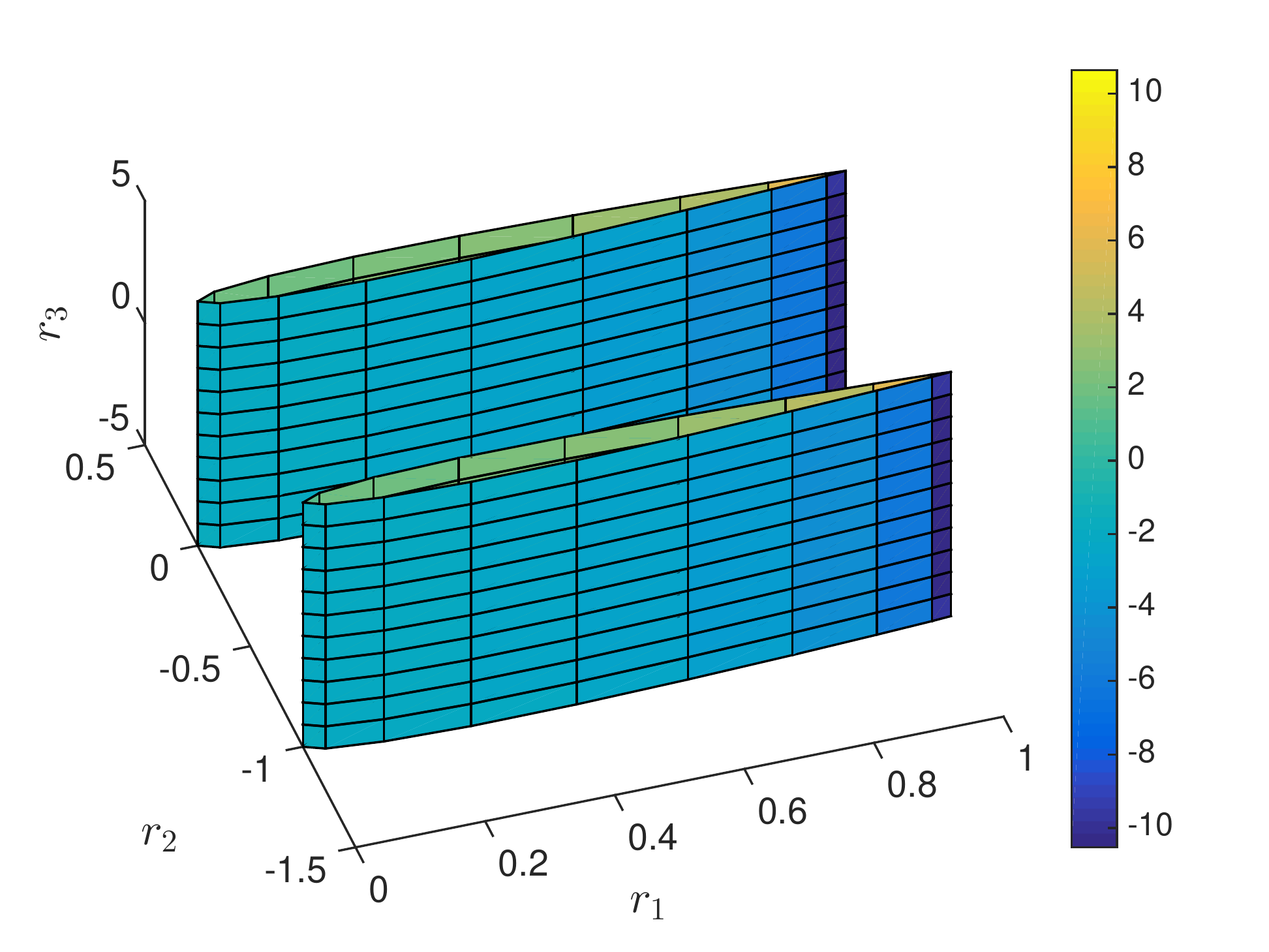}
			\caption{Value of $\mathbf{u}$ on the two interacting wings computed with $\mathcal{T}$.}
			\label{fig:panel_2wings}
		\end{minipage}
		\hfill
		\begin{minipage}[t]{0.47\textwidth}
			\includegraphics[width=\textwidth]{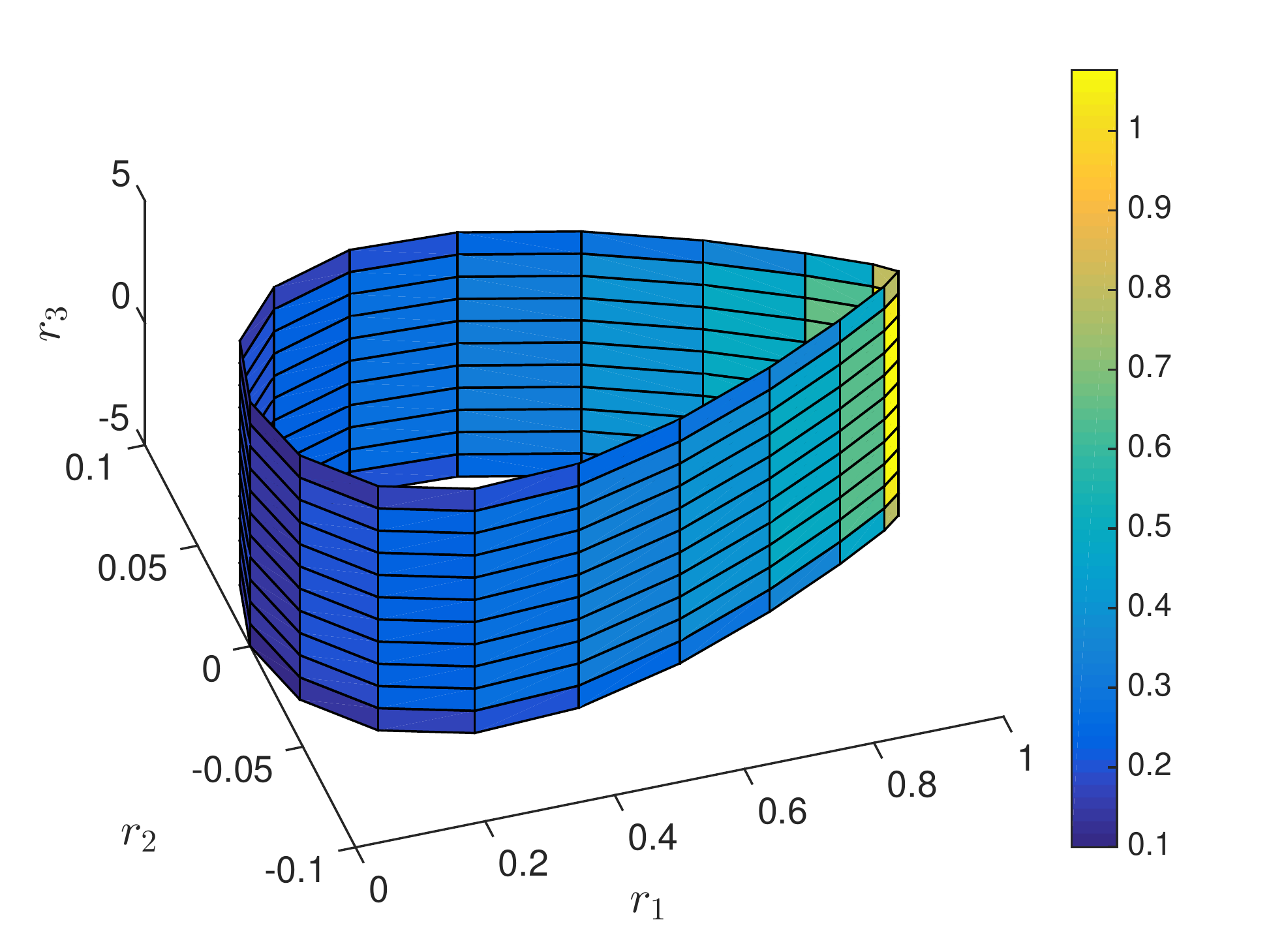}
			\caption{Absolute error in $\mathbf{u}$ on the wing of interest computed with $\mathcal{M}$ (only one wing).}
			\label{fig:panel_1wing}
		\end{minipage}
	\end{figure}

	We consider a problem where an air velocity field $\mathbf{v}$ is induced by two wings moving at velocity $\mathbf{v}_{w}$ (Fig.\ref{fig:panel_2wings}).
	The compressibility of air is neglected, and the curl of the velocity field is assumed to be zero (no vorticity in the flow field).
	Under these assumptions, the velocity vector describing the flow field can be represented as the gradient of a scalar velocity potential $u$: $\mathbf{v} = \nabla u$.
	This leads to the following system of partial differential equations with non-penetration boundary conditions:
	\begin{align}
		\nabla^{2}u &=  0,\\
		\frac{\partial u}{\partial \mathbf{n}} &= \nabla u \cdot \mathbf{n} =  \mathbf{v}_{w} \cdot \mathbf{n},
	\end{align}
	where $\mathbf{n}$ is the outward normal vector at the surface of a wing.

 	The goal of the inverse problem is to estimate the potential $u$ on one of the two wings for a given wing velocity $\mathbf{v}_{w}$ (angle and amplitude). %
	Accurately estimating the potential flow is important because it can be used to estimate the velocity field $\mathbf{v}$ and predict quantities such as drag. 
	Using a panel method \cite{hess1990panel}, it is possible to compute $u$ at the surface of the wing of interest (i.e., without computing $u$ on a mesh as in finite volume methods).
	This leads to the following formulation
	for the fully specified model $T$:
	\begin{align}
		T(\mathbf{u}_{T},\mathbf{q}) = \mathcal{T}\mathbf{u}_{T}-\mathbf{q}=0,
	\end{align}
	where $\mathbf{u}_{T}$ and $\mathbf{q}$ are vectors such that the $j^{th}$ entry $u_{T,j} = u(\mathbf{r}_{j})$ and $q_{j}=  \mathbf{v}_{w} \cdot \mathbf{n}(\mathbf{r}_{j})$, with $\mathbf{r}_{j}$ the centroid location of the $j^{th}$  panel and $\mathbf{n}(\mathbf{r}_{j})$ the corresponding outward normal vector. 
	$\mathcal{T}$ is a $p\times p$ dense matrix induced by a kernel, $\mathcal{K}$, related to the governing equations, where $p$ is the total number of panels used: $\mathcal{T}_{ij} = \mathcal{K}(\mathbf{r}_{i},\mathbf{r}_{j})$.

	In this particular example, $\mathcal{T}\in \mathbb{R}^{396\times 396}$ is the operator corresponding to the two interacting wings ($p=396$).
	The misspecified model $M(\mathbf{u}_{M},\mathbf{q}) = \mathcal{M}\mathbf{u}_{M}-\mathbf{q}=0$, with
	$\mathcal{M}\in\mathbb{R}^{198\times 198}$, is defined only with the panels on the wing of interest ($p=198$) and is thus unable to represent the interaction between the two wings (Fig.\ref{fig:panel_1wing}).
	In realistic systems, one could decide  to neglect such interactions for several reasons, such as the interaction may be believed to be weak, or the dimension reduction of the matrix used for inversion might lead to computational savings.

	We use $N_T=50$ different angles for the velocity $\mathbf{v}_{w}$ (between -45 and 45 degrees) to generate 50 vectors $\mathbf{q}$.
	The fully specified operator is used to compute the associated vectors $\mathbf{u}_{T}$, which are used to generate a data set.
	The 50 data vectors $\mathbf{d}$ are noiseless observations of the state  restricted to the panels corresponding to the wing of interest.
	Those 50 pairs of truncated vectors are used to define a training set $\{D,Q\}$.
	The test set $\{D_{test},Q_{test} \}$ is composed of 100 pairs similarly computed.
	The test set is used to assess the performance of the corrected model, and is not used to construct the correction operator $\mathcal{C}$.
	We apply a modified version of the Frank-Wolfe algorithm \cite{jaggi2013revisiting,hao2014nuclear} to solve the constrained optimization problem Eq.~\ref{eq:reformulated_optim}.

\begin{figure}[!tbp]
	\centering
		\begin{minipage}[t]{0.47\textwidth}
			\includegraphics[width=\textwidth]{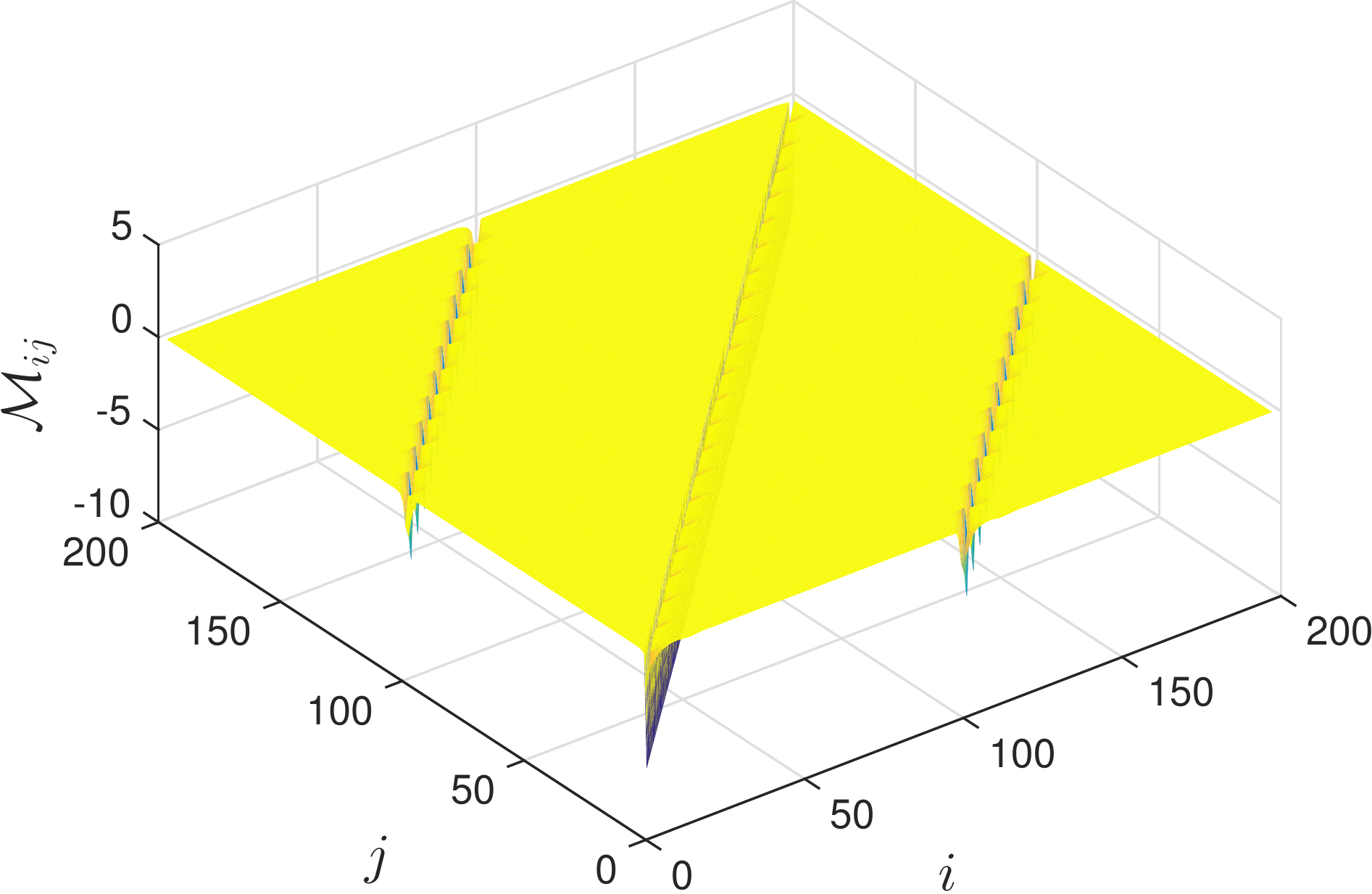}
			\caption{Entries of the misspecified operator $\mathcal{M}$.}
			\label{fig:surfA_panel}
		\end{minipage}
		\hfill
		\begin{minipage}[t]{0.47\textwidth}
			\includegraphics[width=\textwidth]{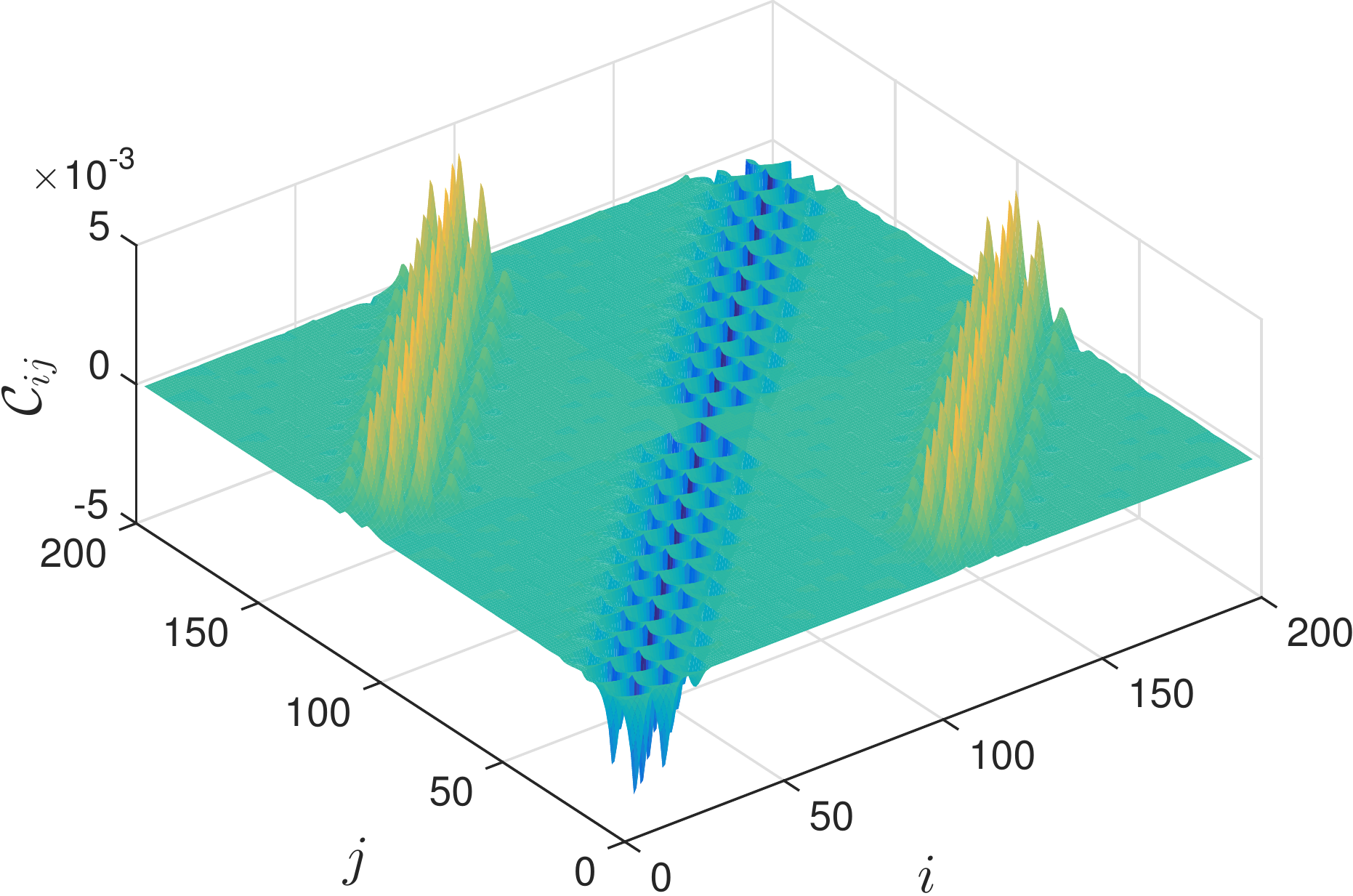}
			\caption{Entries of the correction operator $\mathcal{C}$.}
			\label{fig:surfB_panel}
		\end{minipage}
	\end{figure}

	The model correction problem is solved for several values of the regularization parameter $\lambda$  to infer the additive low-rank, linear, correction operator $\mathcal{C}$.
	Entries of the misspecified operator $\mathcal{M}$, as well as the correction operator $\mathcal{C}$ for $\lambda =10^{7}$ are shown in Fig.~\ref{fig:surfA_panel}--\ref{fig:surfB_panel}.
	The misspecified specified operator exhibits three dominant diagonal structures.
	We notice that the correction $\mathcal{C}$ is characterized by three diagonal structures similarly located but of different values compared to $\mathcal{M}$.
	This structure arises naturally (i.e. without parameterization) from the formulation.

  	\subsection{Proxy for the Condition Number} 
  	\label{sub:proxy_for_condition_number}
  	
\begin{figure}[!tbp]
	\centering
		\begin{minipage}[t]{0.47\textwidth}
			\includegraphics[width=\textwidth]{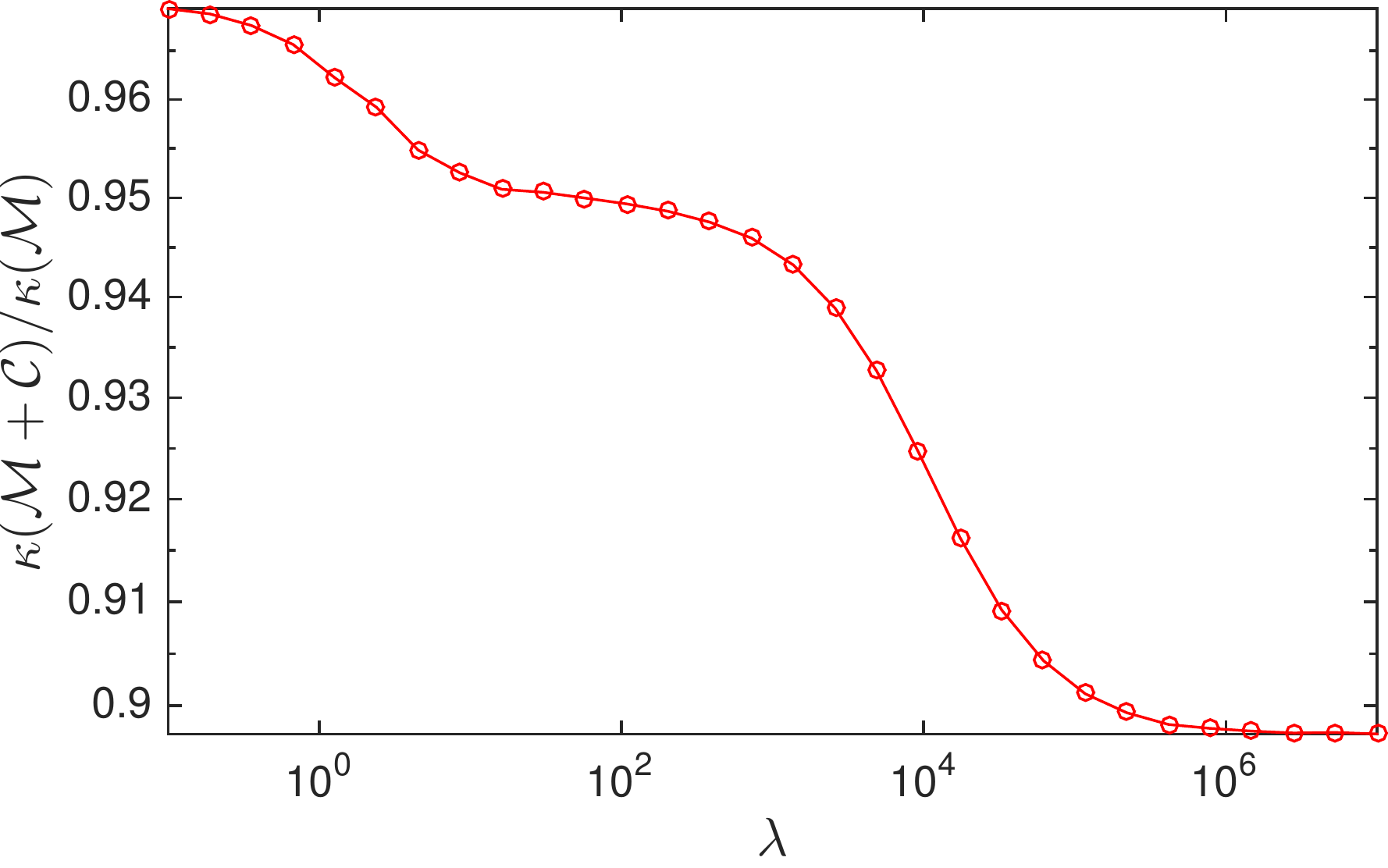}
			\caption{Condition number of the corrected operator as a function of the regularization parameter $\lambda$ (normalized by $\kappa(\mathcal{M})$).}
			\label{fig:cond_panel}
		\end{minipage}
		\hfill
		\begin{minipage}[t]{0.47\textwidth}
			\includegraphics[width=\textwidth]{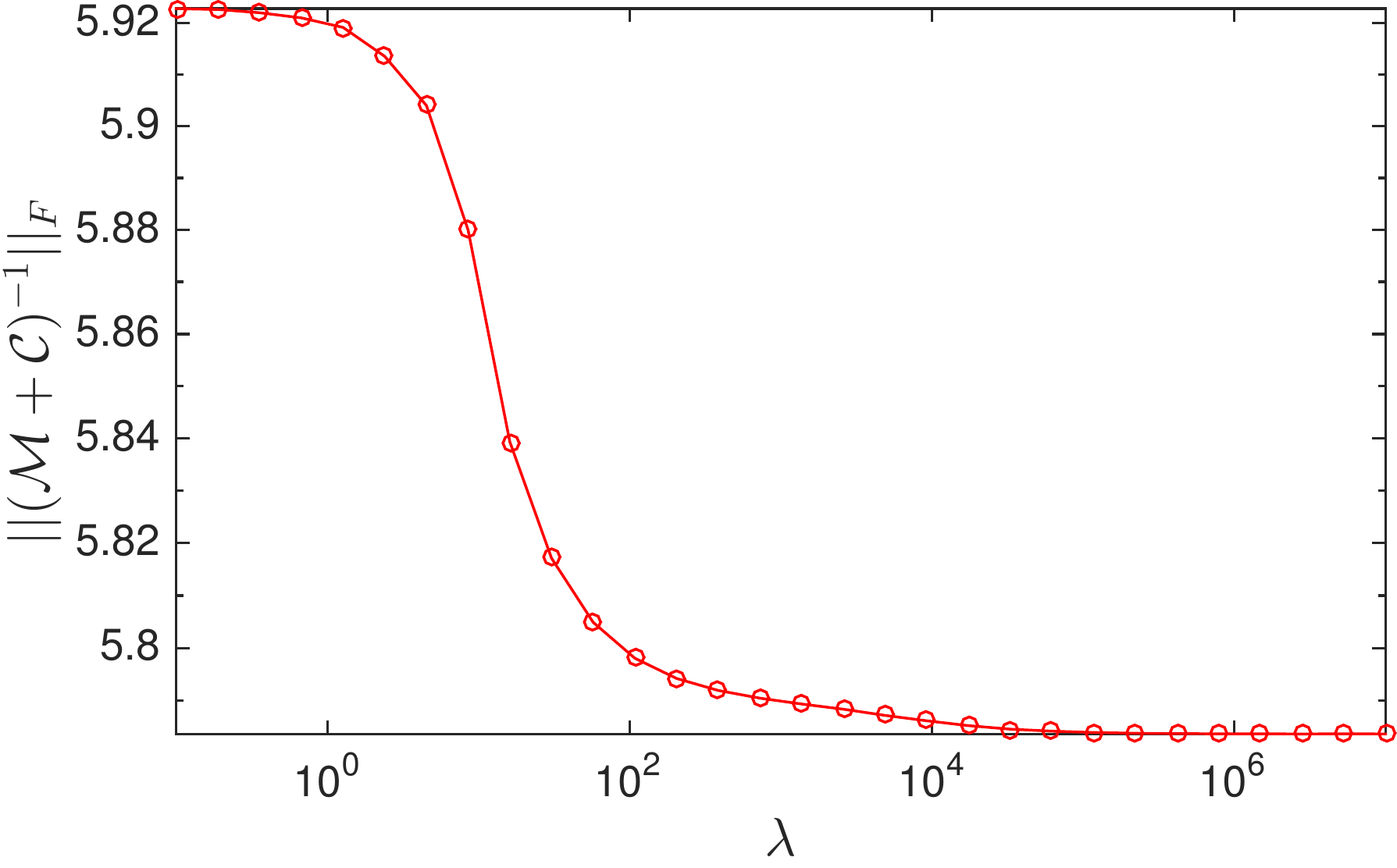}
			\caption{Proxy for the condition number of a corrected operator as a function of the regularization parameter $\lambda$.}
			\label{fig:condProxy_panel} 	
		\end{minipage}
	\end{figure}

	We first examine the condition number of the corrected operator and compare it to the proposed proxy.
	As expected, the condition number $\kappa$ of the corrected operator (Fig.~\ref{fig:cond_panel}) behaves qualitatively similarly to its inverse Frobenius norm proxy (Fig.~\ref{fig:condProxy_panel}) for different values of $\lambda$.
	In particular, for $\lambda\approx1$, both the condition number and the proxy decrease as $\lambda$ increases, for regularization parameters larger than $10^{5}$ both quantities are little affected.

  	\subsection{Error Analysis} 
  	\label{sub:error_analysis}
  	
\begin{figure}[!tbp]
	\centering
		\begin{minipage}[t]{0.47\textwidth}
			\includegraphics[width=\textwidth]{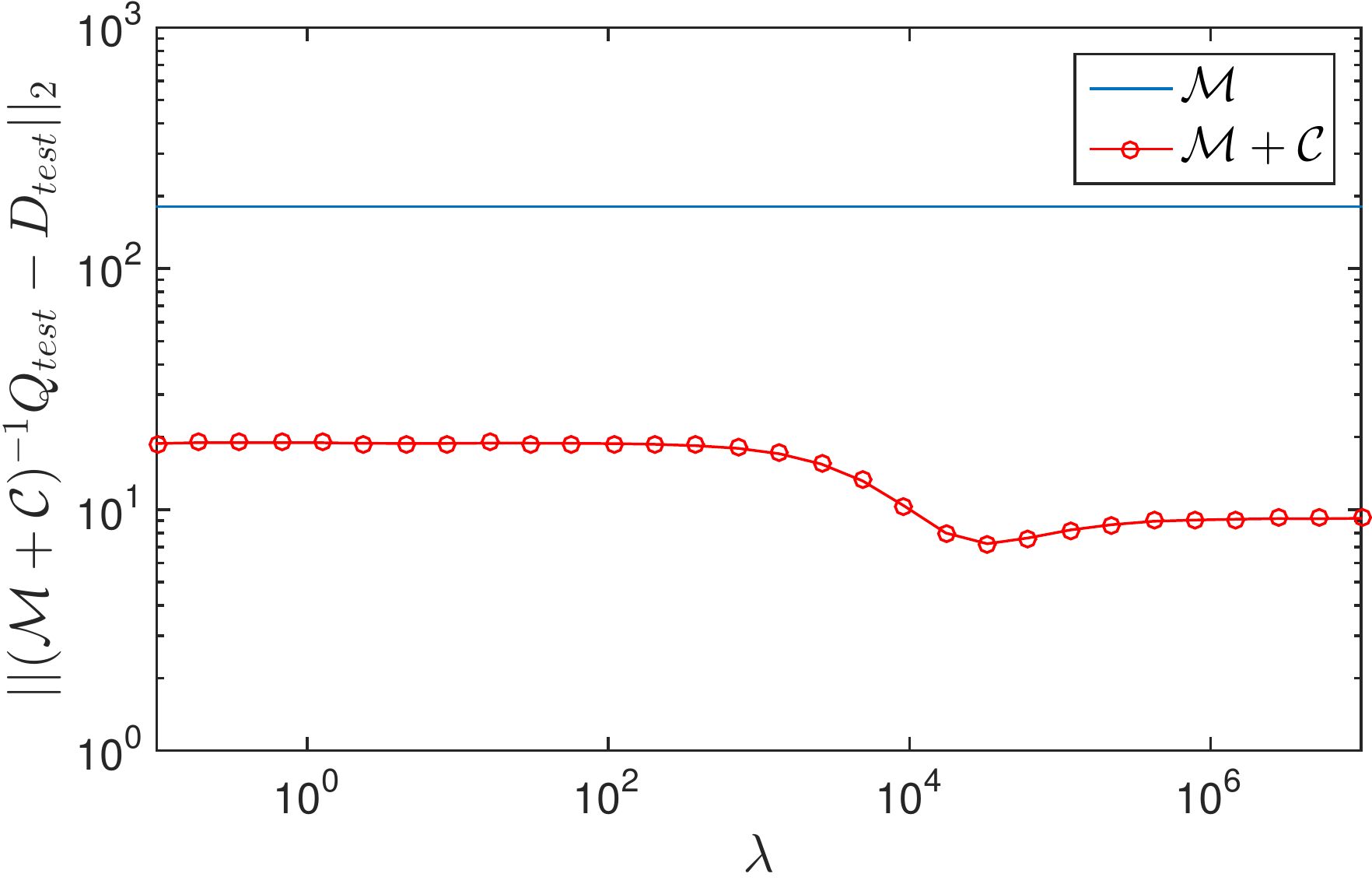}
			\caption{Inverse error on the test set as a function of the regularization parameter $\lambda$.}
			\label{fig:invErr_panel}
		\end{minipage}
		\hfill
		\begin{minipage}[t]{0.47\textwidth}
			\includegraphics[width=\textwidth]{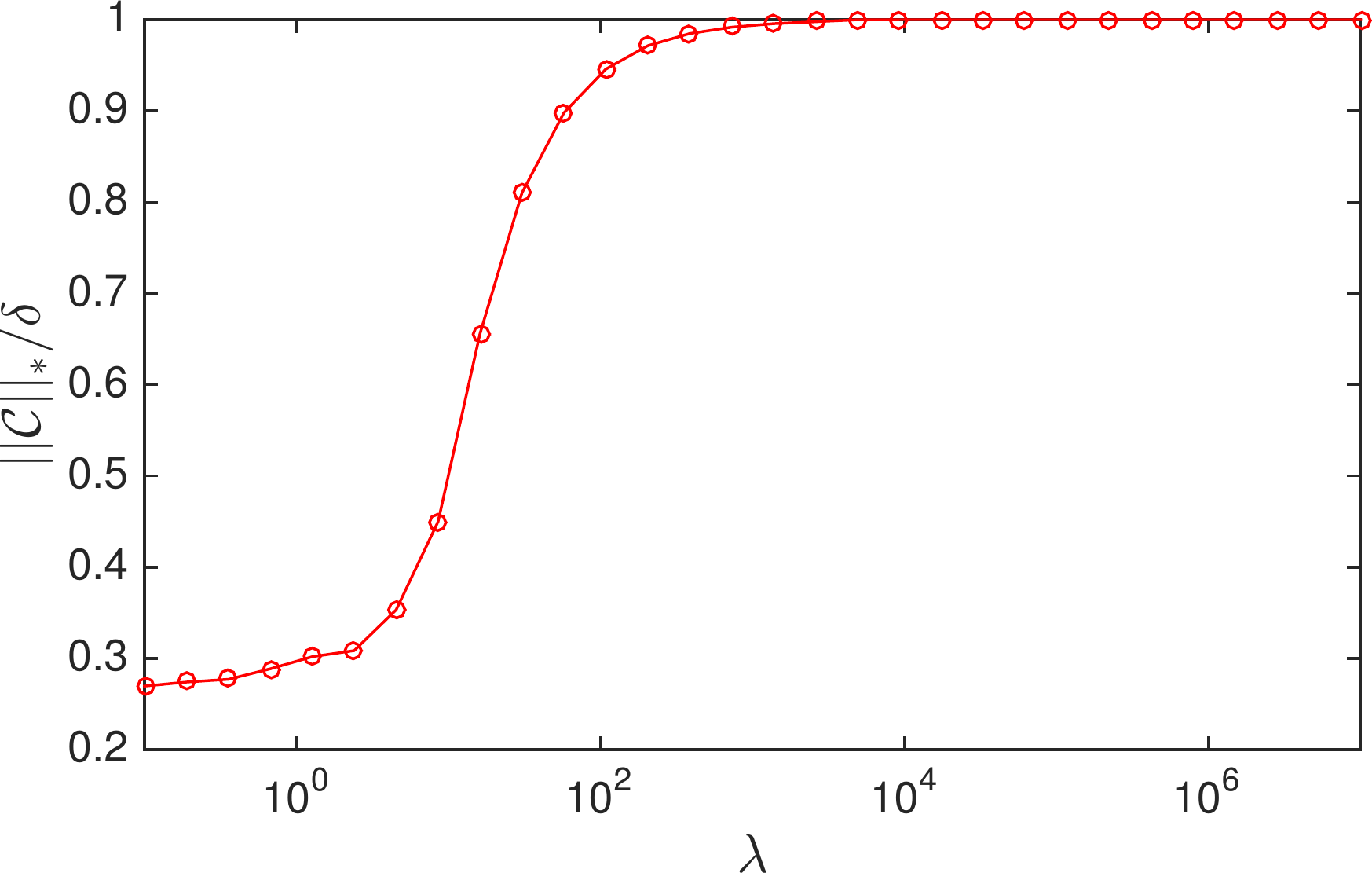}
			\caption{Nuclear norm of the correction operator as a function of the regularization parameter $\lambda$ (normalized by $\delta$).}
			\label{fig:nuclear_norm_panel}
		\end{minipage}
	\end{figure}

	We assess the performance of the proposed method on the test set \sloppy$\{D_{test},Q_{test} \}$. 
	We show the inverse error (Fig.~\ref{fig:invErr_panel}) obtained with corrected operators $\mathcal{M}+\mathcal{C}$ (for different $\lambda$) and compare it  to results computed with $\mathcal{M}$.
	
	We note that, in this case, the corrected models outperform the misspecified model for all regularization parameters $\lambda$.
	For low $\lambda$, the constraint on the nuclear norm (Fig.\ref{fig:nuclear_norm_panel}) is not active. 
	Thus, increasing the regularization parameter decreases the condition number (Fig.\ref{fig:cond_panel}), yet, has little effect on the inverse error (Fig.\ref{fig:invErr_panel}).
	For $\lambda\geq10^{3}$, the constraint becomes active and further reduction on the condition number is observed, with a decrease in inverse error.
	For $\lambda\geq3\times10^{4}$, the condition number decreases further and little emphasis is given to reduction of the inverse error on the training set, which explains the slight increase of the inverse error on the test set.

	Empirically, penalizing the inverse error on a training set to construct a correction operator increases performance on a test set.
	Enforcing a low condition number can further reduce the inverse error for a large range of $\lambda$.
	We note that even for $\lambda\geq 3\times10^{4}$, the inverse error is lower than what would have been obtained by not enforcing the condition number virtue (the limiting case where $\lambda\rightarrow 0$).

\section{Future Perspectives} 
\label{sub:conclusion}

	Our main contribution is an optimization framework in which models seeded from first-principles can be improved
	and augmented via a data-driven approach.
	This construction is cast as an optimization problem with three components:
	a fidelity term to reduce discrepancy with the data, a structure term that characterizes the form of the correction (either implicitly or explicitly), and a virtue that tailors the corrected model to a given end-goal task.
	This framework is quite general and choosing a specific manifestation for these three components is problem dependent.
	
	For the purpose of illustration, we demonstrated our approach using an inversion problem with a low-rank structure for the correction and enforcing a low condition number as virtue.
	We have solved this particular form of the framework using the Frank-Wolfe algorithm and closed-form gradients.	
	This was used to improve a misspecified airfoil panel code lacking interaction between two components.
	Using a training set generated by a fully specified model that modeled the two interacting wings, the correction was able to adjust the model to account for that interaction.

	Ongoing extensions to this work include investigating other forms of structure.
	For example, a sparse correction could be adapted to correct PDE operators while maintaining their sparsity.
	Another extension of the framework is to nonlinear models, which would significantly extend the applicability of the proposed approach to real-world applications.









\bibliographystyle{elsarticle-num-names}
\bibliography{biblio}

\end{document}